\RequirePackage{fix-cm}
\documentclass[twocolumn]{svjour3}          
\smartqed  

\usepackage{subfig}
\usepackage{graphicx}
\usepackage{xcolor}
\usepackage{amsmath}
\usepackage{amsfonts}
\usepackage{textcomp}
\usepackage{cite}

\usepackage[hyphens]{url}
\usepackage{mathbbol}
\usepackage{hyperref}
\usepackage{xspace}

\newcommand{\xmax}{\ensuremath{X_{\mathrm{max}}}\xspace}

\journalname{Computing and Software for Big Science}
\begin{document}

\title{Shared Data and Algorithms for Deep Learning in Fundamental Physics}



\author{
Lisa Benato \and Erik Buhmann \and Martin Erdmann \and Peter Fackeldey \and Jonas Glombitza \and Nikolai Hartmann \and Gregor Kasieczka \and William Korcari \and Thomas Kuhr \and Jan Steinheimer \and Horst St\"ocker \and Tilman Plehn \and Kai Zhou
}


\institute{L.~Benato, E.~Buhmann, G.~Kasieczka, and W.~Korcari \at
Institut f\"ur Experimentalphysik, Universit\"at Hamburg, Germany\\
\email{gregor.kasieczka@uni-hamburg.de}   
\and
J.~Glombitza, M.~Erdmann, and P.~Fackeldey
            \at
III. Physikalisches Institut A,
RWTH Aachen University, Germany
\and
N.~Hartmann and T.~Kuhr \at
Fakultät für Physik,
Ludwig Maximilians University Munich, Germany
\and
J.~Steinheimer, H.~St\"ocker, and K.~Zhou \at
Frankfurt Institute for Advanced Studies (FIAS), Germany
\and
T.~Plehn \at 
Institut für Theoretische Physik, Universität Heidelberg, Germany
}

\date{Received: date / Accepted: date}





\maketitle
\begin{abstract}

We introduce a \textsc{Python} package that provides simply and unified access
to a collection of datasets from fundamental physics research --- including 
particle phy\-sics, astroparticle physics, and hadron- and nuclear phy\-sics --- for supervised machine learning studies. 
The datasets contain hadronic top quarks, cosmic-ray induced air showers, phase transitions in hadronic matter, and generator-level histories.
While public datasets from multiple fundamental physics disciplines already exist, the common interface and provided reference models simplify future work on cross-disciplinary machine learning and transfer learning in fundamental physics.
We discuss the design and structure and line out how additional datasets can be submitted for inclusion.

As showcase application, we present a simple yet flexible graph-based neural network architecture that can easily be applied to a wide range of supervised learning tasks.
We show that our approach reaches performance close to dedicated methods on all datasets. 
To simplify adaptation for various problems, we provide easy-to-follow instructions on how graph-based representations of data structures, relevant for fundamental physics, can be constructed and provide code implementations for several of them. 
Implementations are also provided for our proposed method and all reference algorithms. 

\keywords{deep learning \and fundamental physics \and datasets}
\end{abstract}
\section{Introduction}

Access to high-quality training data is a known key requirement for progress in machine learning and
well-curated open benchmark datasets allow reliable comparisons and promote sustained progress.
Well-known benchmarks and competitions such as MNIST~\cite{lecun-mnisthandwrittendigit-2010}, CIFAR~\cite{krizhevsky2009learning}, and ImageNet~\cite{deng2009imagenet} were pivotal in bringing about the
powerful deep learning models for image processing observed today.
Research in fundamental physics is no different in this regard. Existing datasets and challenges~\cite{pmlr-v42-cowa14,Amrouche:2019wmx,Kasieczka:2019dbj,Kasieczka:2021xcg,Rousseau:2020rnz,calogan,brooijmans2020les,Aarrestad:2021oeb}
are crucial and enable continuous improvements.

A particularly challenging and specific aspect of data science developments in physics is the abundance of needed data representations and accordingly different architecture designs resulting from different experiment designs and theoretical descriptions.
Data might come in the form of high-level observables without
additional structure \cite{pmlr-v42-cowa14},
as rectangular \cite{Almeida:2015jua,deOliveira:2015xxd,Komiske:2016rsd,Kasieczka:2017nvn} or hexagonal \cite{Shilon_2019, pao_dl_2021} images, 
as sequences~\cite{Egan:2017ojy,Erdmann:2019blf} or sequences of images~\cite{PhysRevD.100.011501},
as point clouds of vectors in space~\cite{Komiske:2018cqr,OmanaKuttan:2020brq}, 
as graphs~\cite{Qu:2019gqs}, as data points distributed on non-Euclidean manifolds~\cite{Bister_2021}, or even as hybrid of different sources~\cite{Bols:2020bkb}. 

Of course, it would be possible to work with domain and experiment-specific data representations for each problem.
However, such a strategy would be inefficient and wasteful as different groups and communities need to replicate very similar developments instead of using ready-made solutions.
Therefore, flexible representations and network architectures simultaneously suited for multiple sources of data would be of great potential use and distribute the gains from deep learning for scientific applications in science more readily and widely.

While some common benchmark datasets exist in physics, these usually are domain-specific. Instead, this contribution, together with the provided \textsc{pd4ml} \textsc{Python} package, unifies data across different domains and both from experiment and theory with minimal barriers to access.\footnote{Datasets and models are available at \url{https://github.com/erum-data-idt/pd4ml}}
A key difference to existing datasets and collections is that \textsc{pd4ml} provides a simple interface to download and import all included datasets in a consistent ready-to-use way, without the need for additional pre-processing. We will outline a protocol for extending this collection with additional contributions below.

We focus on supervised learning tasks and include examples of both classification and regression problems.
While weakly-supervised, semi-supervised, and unsupervised learning tasks are of increasing relevance also for physics applications (see e.g. Ref.~\cite{Kasieczka:2021xcg} for
a recent overview), the question of a suitable representation of data can already be answered in a supervised setting.
This approach also benefits from easy-to-define and evaluate performance metrics such as accuracy and the area under the ROC curve (AUC) for classification and the resolution for regression tasks.

Another key feature is that all all datasets come with a reference implementation of a suitable algorithm representing the state-of-the-art for the specific problem and domain. 
Specifically included are: 
\begin{itemize}
\item distinguishing hadronically decaying top-quarks from light quarks or gluon jets in simulated collisions in an LHC experiment (Top Tagging Landscape)~\cite{Kasieczka:2019dbj};
\item identifying interesting events in the Belle~II simulation before detector simulation and reconstruction (Smart Backgrounds)~\cite{james-thesis,CHEP2019-smartbkg};
\item identifying the effects of a non-equilibrium deconfinement phase transition in relativistic nuclear collisions (Spinodal or Not)\cite{Steinheimer:2019iso};
\item assigning the correct nature of phase transitions with additional domain adaptation issues (EOSL)~\cite{Pang:2019int};
\item and reconstructing the shower maximum from the simulated signal patterns recorded by a cosmic-ray observatory (Air Showers).
\end{itemize}

These initial datasets already cover aspects of LHC physics, flavor physics, hadron- and nuclear physics as well as astroparticle physics\footnote{The authors explicitly welcome further contributions of suitable supervised learning problems from fundamental physics and related domains. These can be made by implementing the binding functions following the structure of in \url{https://github.com/erum-data-idt/pd4ml} and creating a pull request. Data can be stored at any place that allows automated downloads, such as the \textsc{Zenodo} platform (\url{https://zenodo.org/})}. The different data volumes and structures ensure that models performing
well on these datasets will generalize well to other datasets in fundamental physics.

Given the diverse nature of data in physics, finding suitable representations and compatible network architectures that simultaneously capture relevant symmetries while providing sufficient expressiveness is a persistent problem.
On one end of the spectrum lie fully-connected networks (FCNs). 
These can be applied to any data, but provide no useful additional symmetries leading to inefficient training and potential overfitting.
On the other end are specific architectures designed for a unique problem. 
These might be simple convolutional operations or even encode specific physical properties~\cite{Butter:2017cot,Erdmann:2018shi,Bogatskiy:2020tje}.
Their main disadvantage is that they need to be fine-tuned to a specific physics challenge.
An intermediate solution --- generic enough to be widely applicable, but with sufficient performance --- could greatly simplify model development for many applications in physics. 
Based on the assembled datasets, we investigate if representing data as graphs has the potential to act as such a \textit{lingua franca}.
Previously, graphs were already successfully explored in a range of fundamental physics 
applications~\cite{Qu:2019gqs,Moreno:2019bmu,Qasim:2019otl,Dreyer:2020brq,Duarte:2020ngm,Heintz:2020soy,Pata:2021oez,Kansal:2020svm,guo2020boosted,alonsomonsalve2020graph,1811770,1808887,Choma:2020cry,1801423,Ju:2020xty,Martinez:2018fwc}.

Graphs are a versatile data structure, consisting of nodes connected by edges. 
Broadly speaking, graph networks process information by passing messages between connected nodes. A detailed description is given in Section~\ref{gnn}.

When applying graph networks to physics data, it is necessary to decide
how to best represent the data as a graph. 
This might be trivial --- for example, if the data are naturally represented as a graph as is the case for shower histories --- or require additional processing steps --- if the data are embedded in a geometrical space and locality in that space can be used to build a graph structure.
Along with each dataset, we present a strategy to build a corresponding graph. 

The remainder of this paper is organized as follows:
We introduce the different physics tasks and datasets in Sec.~\ref{sec:datasets}
and discuss the \textsc{Python} interface in Sec.~\ref{sec:interface}.
An example application of developing and comparing fully-connected and graph-based
common tagging architectures is discussed in Sec.~\ref{sec:models}.
A summary along with a further outlook are given in 
Sec.~\ref{sec:closing} while specific reference architectures for all tasks are summarized 
in Appendix~\ref{sec:reference_models}.

\section{Datasets}

\label{sec:datasets}

An overview of the included datasets is given in Table~\ref{tab:datasets}.
The provided data do not only cover a range of different physics problems but also show the diversity of data regularly encountered in data analysis in fundamental physics.
In the following, additional information on the physics challenge, data generation process, and other details are given for all datasets.

\begin{table*}[]
\centering
\caption{Overview of the provided datasets.\label{tab:datasets}}
\begin{tabular}{lcccc}
\hline
                      & Task           &  Examples & Structure & Dimension \\
                      & & (train/test/validation) & & \\
                      \hline \hline
Top Tagging Landscape & Class. & 1.2M/400k/400k & Four vectors          & 200 particles, 4 features/particle \\
Smart Backgrounds     & Class. & 157k/39k/84k                                            & Decay Graph          & 100 particles, 9 features/particle                               \\
Spinodal or Not       & Class. &   16.3k/4k/8.7k                                             & 2D Histogram              &  20x20 histogram of pion  spectra                                  \\
EoS                   & Class. &         121k/25k/54k                                       & 2D Histogram              & 24x24 histogram of pion  spectra                                  \\
Air Showers     & Regr.     &     56k/30k/14k                                           & 81 1D Traces       & 81 stations, 80 signal bins + timing                                   \\ \hline
\end{tabular}
\end{table*}

\subsection{Top Tagging Landscape}
\label{sec:toptag}

Distinguishing or tagging different types of initial particles using their measured decay products is a very common problem at the LHC and forms the basis for subsequent analysis of collision events.
The identification of hadronically decaying top quarks with a high Lorentz boost is especially relevant as top quarks are a common decay product in many theories predicting physics beyond the Standard Model.

The top tagging reference dataset~\cite{kasieczka_gregor_2019_2603256} was initially produced to allow a fair and consistent comparison of different network architectures developed for this specific problem~\cite{Kasieczka:2019dbj}. 
It is widely used to benchmark classification algorithms in particle physics and its inclusion in the present collection allows a direct comparison of the obtained more general algorithms with the state-of-the-art on this problem.

The dataset consists of light-quarks and gluon (\textit{background}) and hadronically decaying top quarks (\textit{signal}) generated using \textsc{Pythia}~\cite{Sjostrand:2014zea} and detector simulation applied with \textsc{Delphes}~\cite{deFavereau:2013fsa}.
A particle flow algorithm is used for reconstruction, and jets are formed using the Anti-k$_T$ algorithm~\cite{Cacciari:2008gp} with $R=0.8$ implemented in \textsc{FastJet}~\cite{Cacciari:2011ma}. Per event, the 200 constituent four-vectors of the highest transverse momentum jet are stored. Zero-padding is used when fewer constituents are available. Altogether, 1.2M training examples and 400k examples each for testing and validation are provided. Additional details on the dataset and processing are given in Ref.~\cite{Kasieczka:2019dbj}.

\subsection{Smart Backgrounds}
\label{sec:SmartBKG}

Simulation is one of the major computing challenges for experiments like Belle
II that take large amounts of data and require correspondingly large amounts of simulated data.
The full simulation process consists of a
typically fast event generation step, where decay chains from the initially produced particles to stable final state particles are simulated according to measured or predicted decay models, and a more involved detector simulation and reconstruction step.
As it is not known a priori which types of background events are relevant for a particular analysis, the background generation has to include all possible decay chains.
Event selection criteria are applied, to simulated and measured data alike, to increase the significance of signals.
A powerful selection discards a large fraction of background events while keeping the fraction of rejected signal events low.
This is in particular important for searches for rare processes that are highly sensitive to new physics.
The fraction of retained simulated background events can be of the order $10^{-7}$ or even below for analyses at Belle II.
Therefore significant computing resources are wasted to produce simulated background events that are then discarded.

This can potentially be improved by applying filters already after the event generation step, which typically takes only a small fraction of computing resources in the
whole process. In our case, the fraction of time needed for generation is about $0.1$\,\%.
The challenge is to predict which events will pass the final
event selection after detector simulation and reconstruction without running it.
This classification problem was first successfully approached using CNNs in Ref.~\cite{james-thesis} and subsequently improved with graph neural network
approaches \cite{CHEP2019-smartbkg}.

The \textit{Smart Backgrounds} dataset consists of simulated
$e^{+}e^{-}\rightarrow\Upsilon(4S)\rightarrow B^{0}\bar{B}^0$ events with
subsequent decays at the event generation step, created using the EvtGen \cite{Lange:2001uf} event generator in the context of \cite{james-thesis}. The features are the components of the
four-momentum of each particle in the decay chain, their production vertex
positions and time, the particle type, represented by the PDG \cite{Zyla:2020zbs}
identifier, and the index of the mother particle which allows forming a graph of the decay tree. On average 40 particles are generated per event.
Each event in the dataset contains 100 particles, where the features are set
to 0 and the mother particle indices to -1 in events with less than 100
particles in the decay chain. PDG identifiers are mapped to an integer
number between 1 and 506.

The label indicates whether the event passes filtering criteria based on an
event reconstruction using the Full Event Interpretation (FEI) \cite{Keck:2018lcd} algorithm
that tries to reconstruct hadronic $B$ decays from the information
after detector simulation and reconstruction~\cite{Kuhr:2018lps,basf2}.
The FEI selection is shared by many analyses and therefore has a relatively high retention rate of 5\,\%.

\subsection{Spinodal or Not}
\label{sec:spinodal}

The \textit{spinodal} dataset~\cite{steinheimer_jan_2021_5710737} is a simulated dataset, created to identify the effects of a non-equilibrium deconfinement phase transition in relativistic nuclear collisions. The underlying physical question is to understand the non-perturbative interactions of the strong interaction in the high baryon density regime. These questions are closely related to the understanding of the deconfinement mechanism and consequently the dynamical generation of mass in Quantum Chromo Dynamics (QCD). To uncover the properties of dense and hot nuclear matter, the Compressed Baryonic Matter (CBM) \cite{Hohne:2011zza,Senger:2011zza} experiment is under construction at the FAIR (Facility for Anti-proton and Ion Research) at GSI Darmstadt.
At FAIR, heavy nuclei (mostly lead and/or uranium) are accelerated to beam energies of several GeV per nucleon and brought to collision with a similar target at the CBM experiment. In this way, the nuclear matter is simultaneously heated and compressed, and since large nuclei are used, for a very short time, an equilibrated system of QCD matter at several times the nuclear saturation density and temperatures of up to 100~MeV is created. The dynamics that govern this system are determined by the strong interaction, i.e. QCD. Since the dynamic many–body problem of QCD cannot be solved explicitly nor numerically, our understanding of the matter created is based on interpretations of the collected data. This interpretation is done by comparing sophisticated model simulations, either based on relativistic fluid dynamics or microscopic transport simulations, with experimental observations.  
The presented dataset is the result of such a model simulation, based on a fluid dynamical simulation of heavy ion collisions in the presence of a first-order deconfinement phase transition \cite{Steinheimer:2012gc}. In particular, the dataset was created using two distinct scenarios, one where spinodal decomposition occurs and one where it does not. Spinodal decomposition is a well-known effect that describes the dynamics of phase separation and leads to the exponential growth of density fluctuations.  It is now of particular interest how these density fluctuations influence the observable final particle spectra as measured by the CBM experiment. In addition, since even the theoretical background of these fluctuations in the fast-expanding and small collision systems is not well understood, it is even of interest to understand whether all events will show such signals or they only occur on rare occasions. Thus, the application of machine learning methods to possibly uncover the effects of the QCD phase transition measured momentum spectra is of great interest.
The task for this effort is to identify those events which have undergone spinodal decomposition. In addition, for the physical interpretation, it is also important to see what a characteristic spinodal event looks like compared to a non-spinodal event. Finally, it is important to achieve high accuracy to see whether all events, simulated as spinodal events, also can be identified as such or if not every event shows the relevant characteristics.
More information on the physics background and scientific motivation behind this dataset can be found in Ref.~\cite{Steinheimer:2019iso,Steinheimer:2021hoc}. The data included is a coarse-grained --- and hence more difficult --- version of these data.
To create the spinodal classification dataset, 27,000 central collision events of lead on lead are generated at a (typical FAIR/GSI) beam energy of $E_{\mathrm{lab}}=3.5\, A$~GeV, for each scenario: spinodal or not. From each event an 'image' is then generated, containing information on the net baryon density distribution in the transverse spatial $X-Y$ plane. This corresponds then to a 20-by-20 pixel histogram. We renormalized the pictures by their maximum bin value for each event separately, to avoid possible artifacts from one class having a larger density. The histograms are then flattened to a 400-column array-list of events.

\subsection{EoS}
\label{sec:eos}
The EoS dataset is simulated with relativistic hydrodynamics including afterburner hadronic cascade for describing high energy heavy ion collisions. One of the primary motivations for these collision experiments is to understand the QCD phase structure, to which large international efforts have been devoted. The current beam energy scan project at RHIC (BNL) and the forthcoming program at FAIR (GSI) particularly aim at searching for signals and location of the critical end point (CEP) in the QCD phase diagram. 
This CEP separates the crossover transition and the conjectured first-order phase transition from normal hadronic matter to strongly coupled quark-gluon matter. In the laboratory (LHC, RHIC, or future FAIR, NICA), a huge amount of experimental data has been accumulated for heavy ion collisions, making it a fantastic test ground to study these QCD bulk properties. Conventionally, people look into critical fluctuations in experiments to locate the CEP. However, the currently observed signals are too weak to give plausible conclusions. Also, it is rather involved to disentangle many different physical factors (e.g. initial state effects, transport properties, freeze-out physics) in the collision evolution given only the final states observables. Thus, we lack a reliable bridge between the QCD bulk matter properties (EoS) produced in the collisions and the experimental observables. Being different from conventional strategy, we explored another avenue by formulating this task into a classification task well suited to deep learning, see Ref.\cite{Pang:2016vdc,Pang:2019int,Du:2019civ,Du:2020poe,PhysRevD.103.116023}.
The EoS dataset here has the purpose of utilizing deep learning to investigate the nature of the QCD transition happening during nuclei collision, which can be manifested inside the equation of state (EoS) to be employed within the hydrodynamic simulation for the collision. Previously it was demonstrated in Ref.\cite{Pang:2016vdc} that under pure Hydrodynamic evolution (without afterburner hadronic cascade) a deep Convolutional Neural Network (CNN) can identify the nature of the QCD transition encoded in the dynamical evolution with high accuracy provided solely by the pion spectra. This performance was also shown to be robust against other physical factors. Later a further generalization\cite{Du:2019civ} included more realistic simulation compared to experiment was investigated by considering stochastic particalization and UrQMD hadronic cascade followed by hydrodynamics, where the resonance decay effects were taken into account as well in such a hybrid model. The dataset here is the one used in this study. Specifically, the iEBE-VISHNU hybrid model was used to perform event-by-event simulations, with MC-Glauber and also MC-KLN model for the fluctuating initial condition generation. Two different types of EoS are employed in the dynamics: a crossover EoS from lattice-QCD parametrization and the first-order EoS from a Maxwell construction between hadron resonance gas and ideal gas of quark-gluons with a transition temperature at $T_c = 165$~MeV. After the hydrodynamic evolution, the fluid cells are projected into particles (hadrons) using the Cooper-Frye formula, and further, the UrQMD follows for simulating the hydronic cascade. We vary different physical parameters (viscosity, initial condition, switching temperature from hydro to cascade) to generate diverse collision events for the robustness test. This also aims to encourage the network to capture the genuine imprint of the QCD transition onto the final states spectra, instead of being biased by any specific setup of features or uncertain physical properties.
Inside the EoS dataset, the two classes of events were labeled as EOSL and EOSQ, where each event can be viewed as an ``image'' to be the input observable of the deep-learning-based algorithm: the 2-dimensional histogram of pion spectra with 24 transverse momentum bins and 24 azimuthal angle bins.

\subsection{Cosmic-ray induced Air Showers}
\label{sec:airshowers}
When ultra-high-energy cosmic rays (UHECRs) penetrate the Earth’s atmosphere, they induce large particle cascades called extensive air showers.
Using cosmic-ray observatories consisting of ground-based particle detectors, e.g., water-Cherenkov detectors or scintillators, the footprint of air showers can be detected. 
Here, each detector station measures the time-dependent density of shower particles. This results in signal traces, where the shapes encode information on the shower development.
By reconstructing the air-shower characteristics and in particular $\xmax~$--- the depth of shower maximum --- information of the cosmic-ray mass can be obtained.

So far, the reconstruction of \xmax is confined to observations using fluorescence telescopes. However, in contrast to ground-based particle detectors, telescopes have a much smaller duty cycle, as fluorescence observations are only possible in clear and moonless lights. An accurate and extensive measurement of \xmax using ground-based particle detectors could significantly increase the statistics, and thus, give new insights into the cosmic-ray composition.

The air-shower dataset~\cite{glombitza_jonas_2021_5748080} consists of cosmic-ray events and enables the reconstruction of \xmax using a simulated ground-based observatory.
The observatory is placed at the height of $1400$~m above sea level and features a Cartesian detector array with $1500$~m spacing. The design is inspired by the Pierre Auger Observatory~\cite{Auger} and the Telescope Array Project~\cite{TA}.

The used simulation of signals is a fast-simulation approach and utilizes parameterizations and a simplified shower development. It was introduced in Ref.~\cite{ERDMANN201846}.
To decrease memory consumption, simulated air-shower footprints are reduced to a cutout of $9\times9$ stations centered at the station that recorded the largest signal. Thus, the arrangement of detectors can be interpreted as a small 2D image.
A third dimension is given by the time-dependent signal measurement at each detector station.
The signal trace of each station is shortened by selecting the $80$ time steps after the first signals were measured. The width of a one-time step corresponds to $25$~ns. Furthermore, for each detector station, an arrival time is given. It states the arrival time of the first shower particles at each station, which corresponds to the time where the signal traces start.

The goal of this regression task is to predict \xmax from the detector readouts with the best possible resolution, defined as the standard deviation of the distribution given by the difference between the predictions and the actual values of \xmax.

\section{Python Interface}

\label{sec:interface}

To make the datasets conveniently accessible to a wide range of users, we provide a \textsc{Python} interface to make their use as simple as possible. With a few lines of code, it is possible to load any of the five datasets, perform dataset-specific pre-processing, and create a dataset-specific adjacency matrix for graph neural network algorithms.
The package is easily extendable by providing additional wrappers to allow the inclusion of other datasets in the same environment.

The essential function is \verb!load! which allows loading of the training and testing datasets. The dataset features \textbf{X} as well as the labels \textbf{y} are returned as \textsc{NumPy}~\cite{harris2020array}
arrays.
We can see below an example of code that loads the train and the test set of the Spinodal dataset:
\begin{verbatim}
    from pd4ml import Spinodal  
    X_train, y_train  = Spinodal.load(`train')
    X_test, y_test    = Spinodal.load(`test')

\end{verbatim}
If the dataset is not found at the provided \verb!path!, it is automatically downloaded\footnote{The choice of storage location lies with the dataset providers.}. The option to force the download is provided and an \textsc{MD5} check is implemented.

The package provides an additional loading method called \verb!load_data!. In this case, the loaded data undergo a pre-processing routine implemented by the dataset providers. One can require the data to be loaded in the form of a \textit{graph} by setting the appropriate value to the \verb!graph! argument.
The returned adjacency matrices are constructed in the same way described in Sec.~\ref{sec:adj_matrix}.
An example would be:

\begin{verbatim}

    X, y = Spinodal.load_data(`train',
                              path = `.',
                              graph = True)
                              
\end{verbatim}
The dataset description can be printed out by executing the following command:
\begin{verbatim}
    Spinodal.print_description()
\end{verbatim}
The output contains both technical information on the dataset (like the number of events) and references to the papers providing the physics context and the in-detail description of the reference models.

Additionally, we made available in the
repository\footnote{\url{https://github.com/erum-data-idt/pd4ml}} the 
\textsc{TensorFlow}~2.3~\cite{tensorflow2015-whitepaper} implementation of  the reference models described in Sec.~\ref{sec:reference_models} and the common models presented in Sec.~\ref{sec:models}.

\section{Example Application}

\label{sec:models}

To demonstrate the potential added value of such a common interface for transfer learning and similar tasks, we consider two common architectures: FCNs (Sec.~\ref{fcn}) and GraphNets (Sec.~\ref{gnn}).
In both cases, the same model (the only difference being a different number of input layers) is trained on all datasets. 

Additionally, a reference model, against which the performance can be compared, is provided for each data\-set. These are described in Sec.~\ref{sec:reference_models}.
In the case of well-established datasets (such as the Top Tagging Landscape task), state-of-the-art models are used. For novel datasets, an implementation is provided by the dataset providers as well.
\footnote{Contributions of additional algorithm implementations via github pull-requests are possible as well.}

\begin{figure}[tb]
  \centering
     \includegraphics[width=0.37\textwidth]{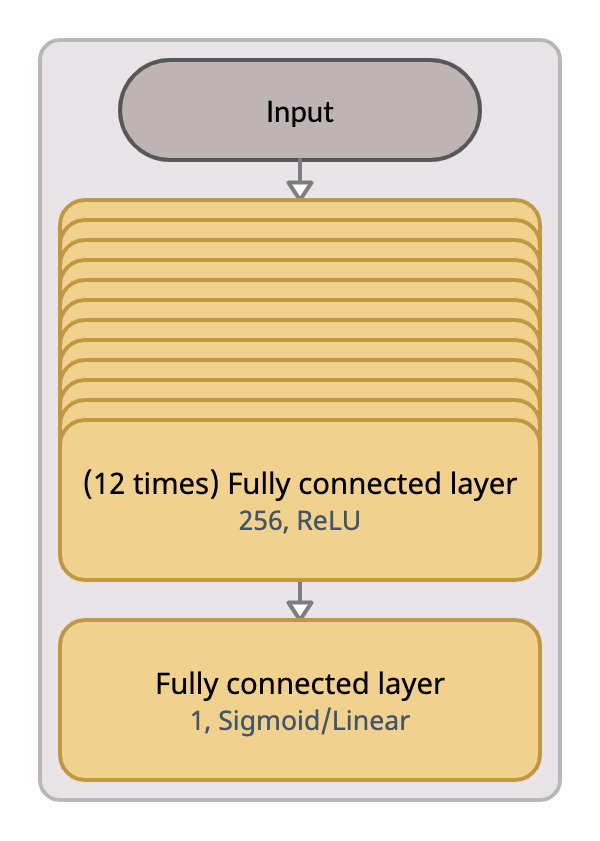}
     \caption{Schematic representation of the FCN. Additional details are provided in the text.
     For classification tasks Sigmoid is used as loss in the final layer, for regression a linear activation is used instead.
     }
     \label{fig:fcn}
\end{figure}

\subsection{Fully-Connected Network}

\label{fcn}

Networks based on the FCN architecture are a useful and versatile baseline as
they require no assumptions about the structure of the data.
For this reason, we construct an FCN architecture to be trained and evaluated on all datasets described in Sec.~\ref{sec:datasets}.
The network has a depth of 12 hidden layers each with a size of 256 nodes and the ReLU activation function.\footnote{No extensive tuning of these hyperparameters was performed but we verified that changes in the number of layers, batch size, and learning rate did not significantly alter the results.}
The output layer, as well as the loss function, are dependent on the task of the network:
\begin{itemize}
    \item Classification task: Sigmoid output and Binary Cross Entropy loss;
    \item Regression task: Linear output and Mean Squared Error loss.
\end{itemize}

The input data are split into batches of 256 examples and the training is performed across 300 epochs regulated by early stopping which terminates the training when no improvement of the validation loss is observed for more than 15 epochs to prevent overfitting effects. 
We train using the the Adam~\cite{kingma2017adam} algorithm with an initial learning rate of 0.001. The learning rate is reduced by a factor of 10 if no improvement of the validation loss is observed within 8 epochs from the last best score.

\subsection{Graph-based Network}

\begin{figure*}[tb]
  \centering
     \subfloat[Blocks][Constituent blocks of the GraphNet]
     {\includegraphics[width=0.405\textwidth]{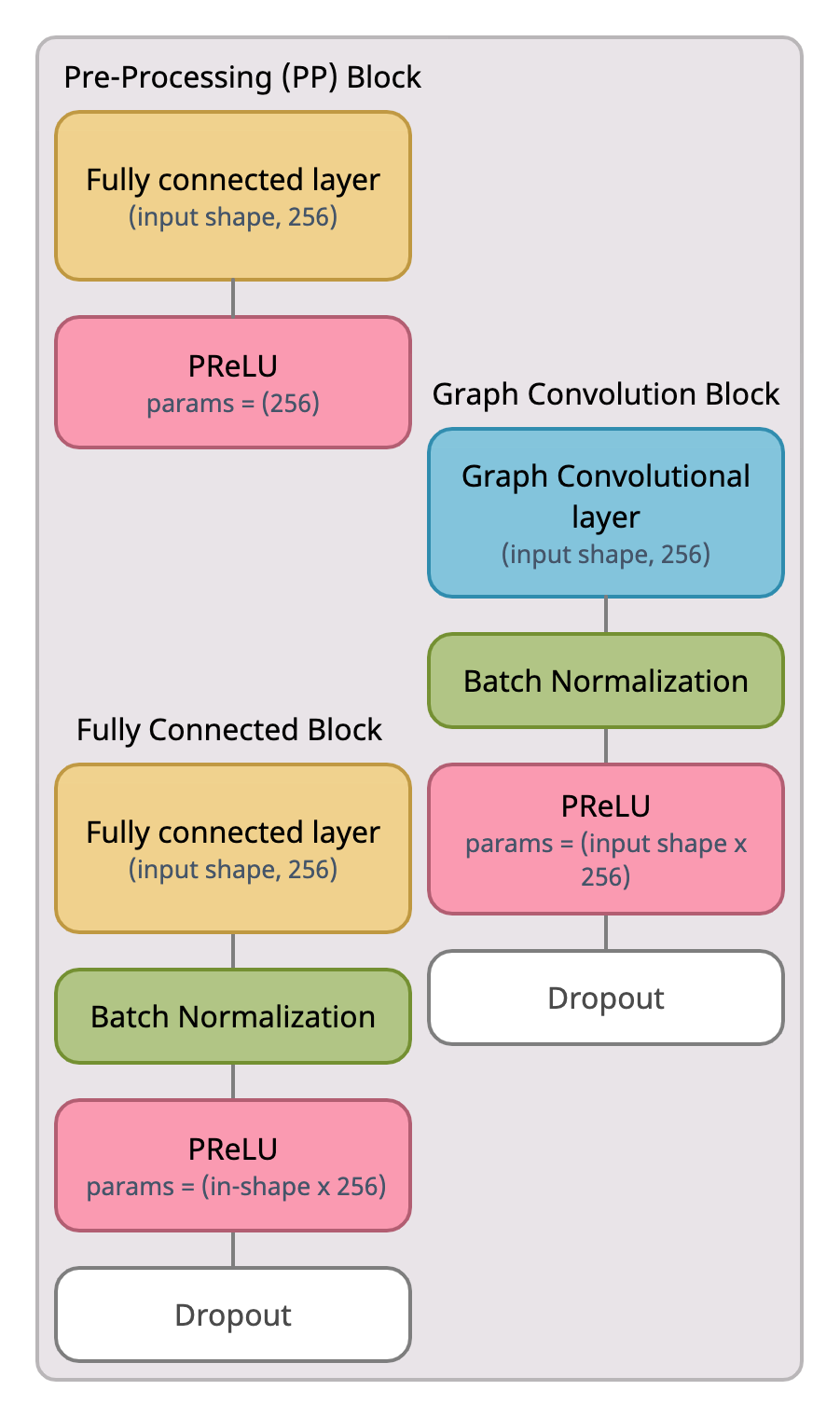}\label{fig:Blocks}} \quad
     \hspace{2cm}
     \subfloat[GraphNet][GraphNet]
     {\includegraphics[width=0.4\textwidth]{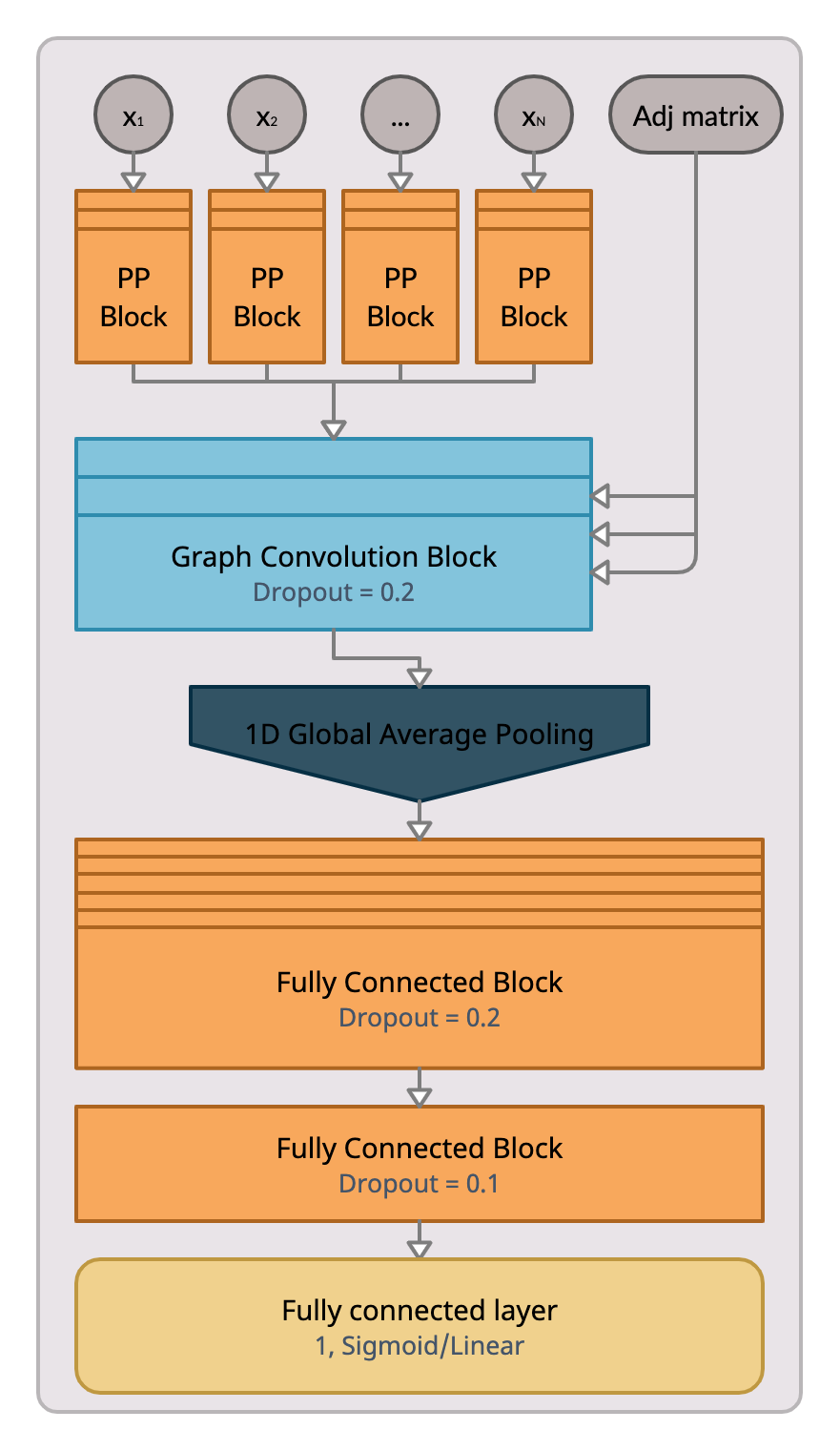}\label{fig:graphNet_struct}}
     \caption{Schematic representation of the GraphNet model. Additional details are provided in the text.}
     \label{fig:graphNet}
\end{figure*}

\label{gnn}

Many types of data can either be naturally represented as a graph or transformed into a graph without loss of information.
This is also the case for data in fundamental physics where measurement signals can be represented as nodes in a graph and their relations or the relation of the measurement devices as the edges. 
This additional information about distances and connectivity between data points is something a simple FCN cannot process.


In the following, we consider undirected graphs with $N$ nodes --- meaning that an edge between node $i$ and $j$ is equivalent to an edge between $j$ and $i$. 
We further assign a vector of features $n_i \in \mathbb{R}^M$
(where $M$ is the dimension of the feature space) to each node. 
Only unweighted graphs are considered for simplicity, although assigning features to the edges is possible as well.
In this case, we can describe the graph representing one input example by two matrices:
the feature matrix $X \in \mathbb{R}^{M \times N}$
and the adjacency matrix $A \in {0,1}^{N \times N}$.
An entry of $1$ in the row $i$ and column $j$ of the adjacency matrix denotes that there is a connection from node $i$ to node $j$. Zeros imply unconnected nodes.
A dataset then consists of many such graphs, in general, one feature matrix and one adjacency matrix per data point.

In principle, most types of data can be turned into a graph without
losing information. We have outlined a few examples in this paper. 
For example, data represented as an image can easily be turned into a graph structure by building an adjacency matrix based on each pixels' neighboring pixel (see Sec.~\ref{sec:adj_matrix}). If the absolute pixel position is relevant, it can be added as per-node feature.

The architecture of our common GraphNet implementation is very similar to the Smart Backgrounds reference model in Sec.~\ref{sec:ref_smartbkg}. The input data are split into batches of 32 elements and the training is performed across 400 epochs regulated by an \textit{early stopping} with patience of 50 epochs. To optimize the training the Adam algorithm with an initial learning rate of 0.001 is used that is then reduced by a factor of 10 if no improvement of the validation loss is observed within 8 epochs from the last best score.

The main difference in comparison to the FCN is given by the addition of an adjacency matrix that is input to the graph convolutional layers \cite{GClayer_kipf2017semisupervised}.
The architecture consists of three fully-connected layers per node with weigth-sharing across nodes, followed by three graph convolutional layers, followed by a 1D global average pooling operation, followed by three fully-connected layers, and an output layer.
Starting from the graph convolutional layers, after each layer batch normalization is applied as well as dropout \cite{jmlr_drpt2014} (fraction of 0.2 for all layers except the last one where the fraction is 0.1).
All layers employ 256 trainable nodes and PReLU \cite{DBLP:journals/corr/HeZR015} activation functions, except the output layer using a sigmoid or a linear activation function depending on the task (classification or regression).
A schematic representation is given in Fig.~\ref{fig:graphNet}.
The same preprocessing steps are performed as for the FCN. Additionally, a dataset-specific step is required to build the adjacency matrices as explained in the following section.

\subsection{Pre-Processing and Adjacency Matrix}
\label{sec:adj_matrix}

For each dataset, basic data preprocessing is performed. The dataset-specific preprocessing is closely inspired by its counterpart in the reference models. 
The Top Tagging Landscape dataset is transformed from four vectors to four hadronic coordinates, more specifically into the logarithm of the transverse energy (log(p$\mathrm{_T}$)), the logarithm of the energy (log(E)), the relative pseudorapidity ($\mathrm{\Delta\eta}$), and the relative $\mathrm{\Delta\phi}$ angle.
For the SmartBKG dataset, the Particle ID information is one-hot encoded, while not changing the other features.
The EoS dataset is standardized and the air-shower dataset is preprocessed by taking the logarithm of the filled signal bins and by normalizing the timing.
No preprocessing is performed for the Spinodal or Not dataset.

\begin{table*}[tbh]
\centering
\caption{Summary how different structures can be represented as graphs\label{tab:graphify}}
\begin{tabular}{lll}
\hline
Dataset           &  Structure & Graph building \\
                      \hline \hline
Top Tagging Landscape &  Four vectors          & $k$-nearest neighbour clustering \\
Smart Backgrounds     &  Decay Graph          & (not needed)                       \\
Spinodal or Not       & 2D Histogram              &  adjacent pixels\\
EoS                   & 2D Histogram              &  adjacent pixels \\
Air Showers     & 81 1D Traces       & geometric relation of detector stations \\ \hline
\end{tabular}
\end{table*}

Each adjacency matrix has been designed to suit the specific dataset characteristics. For the Top Tagging Landscape dataset (Sec.~\ref{sec:toptag}) the adjacency matrix has been built performing a $k$ nearest neighbor clustering using the information of the jet constituents with $k = 7$. In the case of the SmartBKG dataset, the generator-level particles have been used in combination with the mother-daughter relation. Since the EoS and the Spinodal data come in the form of 2D histograms, we exploit this image-like structure and build the adjacency matrix counting as neighbors the eight (three at corners, five at edges) adjacent bins to each ``pixel''. In the air-shower dataset the detectors are arranged in a nine-by-nine rectangular grid, which allows the construction of the adjacency matrix implementing the ``eight adjacent bin'' technique as well. A summary of these graph-building methods is given in Table~\ref{tab:graphify}. 

%

\subsection{Performance Evaluation}
\label{sec:performance}

The performance of all the models has been evaluated by five trainings using the same data but different random initialization.
Then the mean and standard deviation of the obtained scores are calculated. The \textit{accuracy} (fraction of correctly classified data points) and the \textit{area under curve} (AUC) are given for the classification task, while for the regression task we use the \textit{resolution} as described in Sec.~\ref{sec:airshowers}.

\begin{figure*}
  \centering
     \includegraphics[width=0.8 \textwidth]{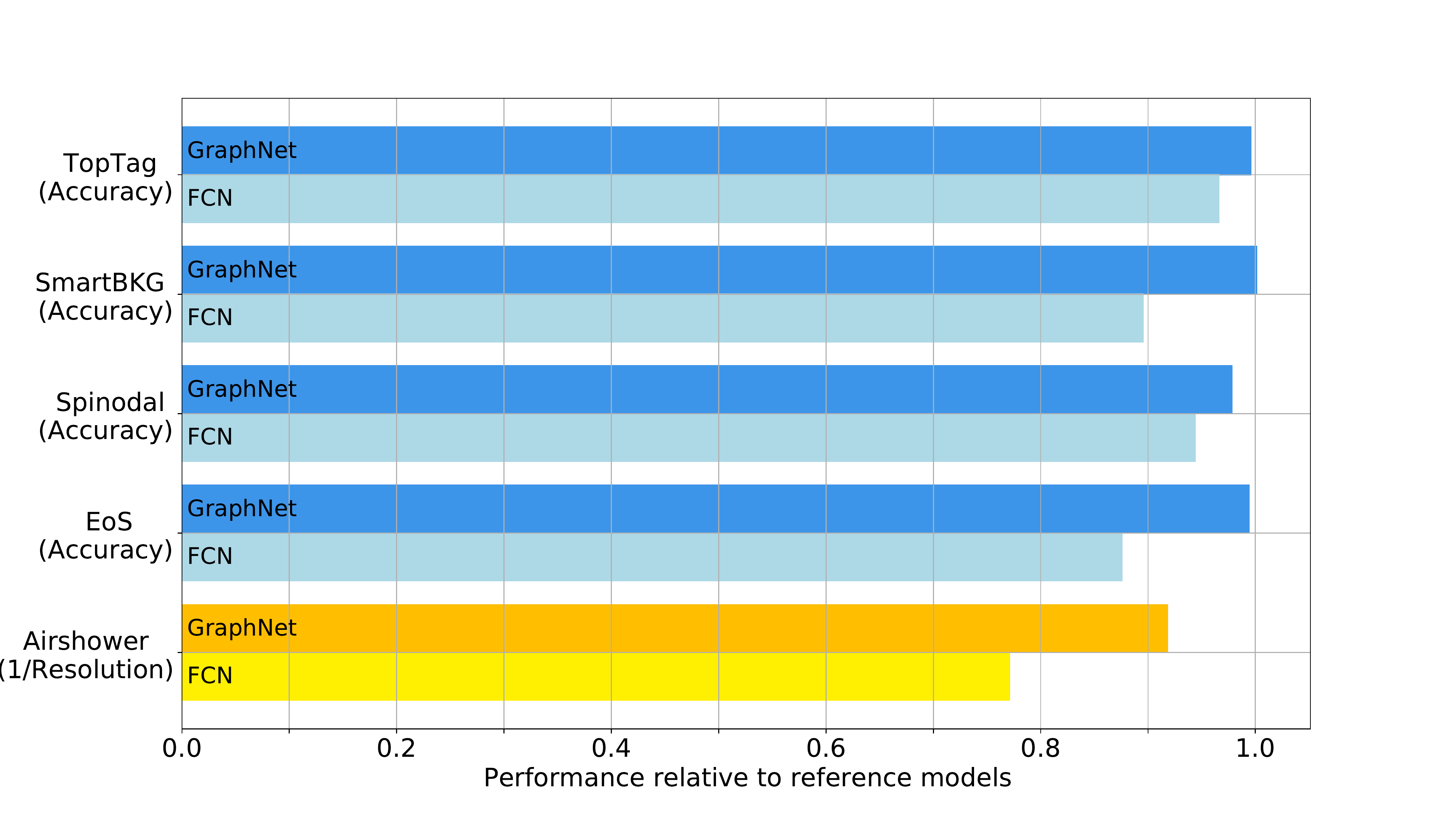}
     \caption{Performance of the GraphNets and the FCNs relative to the reference models for the different datasets.}
     \label{fig:performance}
\end{figure*}

An overview of the results is given in Tab.~\ref{tab:performance_ACC} (accuracy) and Tab.~\ref{tab:performance_AUC} (AUC).
A graphical representation of the accuracy relative to the accuracy achieved by the reference model is given in Fig.~\ref{fig:performance}.
The GraphNet achieves equal or similar performance as the reference model on all classification tasks.
The largest gap is observed for the Spinodal dataset where the graph network underperforms the reference model by 2\,$\%$ in AUC (1\,$\%$ in accuracy).
The performance difference to the FCN is much larger --- here the largest difference is 12\,$\%$ for the EoS dataset.
Regression performance on the air-shower dataset (Tab.~\ref{tab:performance_regression}) shows a larger drop in performance with a ten percent worse resolution for the GraphNet and 30\,\% for the FCN.




\begin{table}[h]
\begin{center}
\label{tab:performance}
\caption{Accuracy scores of the different models. Provided are the mean value and its standard deviation of five independent trainings on the same data.\label{tab:performance_ACC}}

\begin{tabular}{lccc}

\hline
 & Reference & GraphNet& FCN \\
 \hline \hline
TopTag & 0.940 $\pm$ 0.001 &  0.935 $\pm$ 0.001 & 0.908 $\pm$ 0.001 \\
 \hline
SmartBkg & 0.823 $\pm$ 0.001 & 0.824 $\pm$ 0.001 & 0.737 $\pm$ 0.002\\
 \hline
Spinodal &0.873 $\pm$ 0.004 & 0.854 $\pm$ 0.004  & 0.824 $\pm$ 0.001 \\
\hline
EoS &0.691 $\pm$ 0.005 & 0.687 $\pm$ 0.005 & 0.605 $\pm$ 0.019 \\
\hline
\end{tabular}
\end{center}
\end{table}

\begin{table}[h]
\begin{center}
\caption{AUC scores of the different models. Provided are the mean value and its standard deviation of five independent trainings on the same data.\label{tab:performance_AUC}}

\begin{tabular}{llll}

\hline
 & Reference & GraphNet& FCN \\
 \hline \hline
TopTag & 0.986 $\pm$ 0.001 &  0.983 $\pm$ 0.001 & 0.966 $\pm$ 0.002 \\
 \hline
SmartBkg & 0.906 $\pm$ 0.009 & 0.903 $\pm$ 0.001 & 0.811 $\pm$ 0.001\\
 \hline
Spinodal &0.925 $\pm$ 0.005  & 0.916 $\pm$ 0.005  & 0.883 $\pm$ 0.001  \\
\hline
EoS &0.788 $\pm$ 0.005 & 0.766 $\pm$ 0.005 & 0.739 $\pm$ 0.008 \\
\hline
\end{tabular}
\end{center}
\end{table}

\begin{table} [!h]
\begin{center}
\caption{MSE and resolution values measured on the air-shower dataset. Provided are the mean value and its standard deviation of five independent trainings on the same data.\label{tab:performance_regression}}
\begin{tabular}{lll}
\hline
Air Shower    & MSE & Resolution\\
    \hline \hline
  Ref. Model  & 1000 $\pm$ 52 &  31.32 $\pm$ 0.75\\
    \hline
  GraphNet  & 1185 $\pm$ 26 & 34.12 $\pm$ 0.47  \\
    \hline
  FCN  & 1661 $\pm$ 19 &  40.63 $\pm$ 0.40 \\
    \hline
\end{tabular}

\end{center}
\end{table}



%

\section{Outlook and Conclusions}

With the increasing use of machine learning in fundamental physics, a corresponding need for common datasets and algorithms that can be used across various scientific tasks arises. The collected datasets highlight various facets of this challenge and provide data from very different problems and communities in a unified and simple-to-use setup with the extension to more data and learning tasks planned for the future.

The obtained results show that a common graph-based architecture can achieve (close to) state-of-the-art performance with shared hyperparameters across all tasks. This is encouraging and the proposed architecture\footnote{implementation available at \url{https://github.com/erum-data-idt/pd4ml}} can
serve as a solid starting point --- or perhaps even algorithm of choice without further optimization --- for machine learning tasks in fundamental physics. Similarly, the discussed methods to build graphs for common data structures serve as a template for similar applications. 

However, the observed difference in performance for cosmic-ray induced air-shower data implies that further improvements, especially for highly complex data, are needed. Similarly, there is no reason why the reference models should be the upper limit and a sufficiently advanced common model might outperform the baseline on all tasks. Innovations in architectures are needed to overcome this challenge.

These results are just the first steps in taking machine learning in our fields from innovative developments to industrial-style production. A common baseline architecture makes automated training and optimization on user-provided datasets without further manual intervention possible. 
Such a service could greatly reduce the turn-around and development time for many standard applications of machine learning in our field. Of course, exotic, novel, or otherwise unique tasks will still require dedicated efforts. Thinking further, one could also consider using similarities in the 
structure of data representing physical measurements and further improve the performance via transfer learning across datasets.
\label{sec:closing}

\begin{acknowledgements}
This work would not have been possible without the
collaborative project IDT-UM (\textit{Innovative Digitale Technologien zur Erforschung von Universum und Materie}) funded by the German Federal Ministry of Education and Research BMBF and overseen by Projekttr\"ager DESY. We thank 
all our involved colleagues for the excellent collaboration and stimulating
scientific environment.
LB, EB, GK and WK acknowledge the support of the Deutsche Forschungsgemeinschaft (DFG, German Re\-search Foundation) 
under Germany’s Excellence Strategy – EXC 2121  ``Quantum Universe" – 390833306. 
NH and TK acknowledge the support of the Deutsche Forschungsgemeinschaft
under Germany’s Excellence Strategy – EXC 2094  ``ORIGINS" – 390783311. 
We thank the Belle II data production team for producing the Smart Backgrounds dataset used in this study.

\end{acknowledgements}

\section*{Conflict of interest}
The authors declare that they have no conflict of interest.

\newpage
\begin{appendix}

\section{Reference Models}

\label{sec:reference_models}

In the following, a reference model is provided for each dataset. These reference models represent an algorithm specifically designed and optimized for the problem at hand. They serve as a benchmark to judge the performance of the proposed dataset-independent algorithms.

\subsection{Top Tagging Landscape}
\label{sec:ref_toptag}

The ParticleNet~\cite{ParticleNet_Qu_2020} algorithm is used as reference for
the Top Tagging Landscape dataset.
This architecture was one of the best models in a comparison study performed earlier on this dataset~\cite{Kasieczka:2019dbj}. 
The graph convolutional network is constructed by viewing
the input data as point clouds, i.e. each jet is considered an unordered set of particles. To build these point clouds, the particles in each jet are ordered by transverse momentum and zero-padded to up to 100 particles per jet. From the particles 4-momenta seven input features for the network are computed: the logarithm of the transverse energy (log(p$\mathrm{_T}$)), the logarithm of the energy (log(E)), the relative pseudorapidity ($\mathrm{\Delta\eta}$), the relative $\mathrm{\Delta\phi}$ angle, the logarithm of the particle's p$\mathrm{_T}$ relative to the jet p$\mathrm{_T}$ (log($\frac{\mathrm{p}\mathrm{_T}}{\mathrm{p{_T}(jet)}}$)), the logarithm of the particle's energy relative to the jet energy (log$\mathrm{\frac{E}{E(jet)}}$), and the angular sepeartion between the particle and the jet axis ($\Delta R = \sqrt{(\Delta \eta)^2 + (\Delta \phi)^2}$). The relative angles are calculated with respect to the jet axis. 
Based on these inputs, a graph is built for every jet. The edges of this graph are found by a $k$ nearest neighbor search of the $k=7$ closest particles. 
The architecture itself is build of three EdgeConv layers~\cite{EdgeConv_wang2018dynamic}, followed by a global pooling operation over all the particles to preserve permutation invariance, followed by two fully-connected layers. 

\subsection{Smart Backgrounds}
\label{sec:ref_smartbkg}

The graph structure of the event decay tree is an ideal input for graph neural networks. A structure based on graph convolutional layers~\cite{GClayer_kipf2017semisupervised} provides high classification performance.
The PDG identifiers are passed through an embedding layer to produce an
8-dimensional embedding space which is subsequently concatenated with the
remaining 8-dimensional feature vector. These combined particle features are
fed to 3 fully-connected layers, where the weights are shared across all
particles, followed by the ReLU activation function. The next transformation
consists of 3 graph convolutional layers, where the adjacency matrix of the
decay graph is constructed from the indices of mother particles. Furthermore, it
is symmetrized to provide mother-daughter in addition to daughter-mother
relationships. By taking the average of the resulting vector along the particle
dimension it is reduced to event-level quantities which are finally transformed
by 3 more fully-connected layers, followed by the ReLU activation function. A
final fully-connected layer, followed by the sigmoid activation function
provides the classification score. All hidden layers, including the graph convolutional layers, contain 128 units.

\subsection{Spinodal or Not}
\label{sec:ref_spinodal}

Since the simulated output for the data corresponds to a 20x20 histogram, a natural choice for the network structure is that of a convolutional neural network. In this particular case we employ three convolutional layers with intermediate pooling layers, followed by a single hidden fully-connected layer and the output layer. The complete structure of the network is shown in Ref.\cite{Steinheimer:2019iso}.

\subsection{EoS}
\label{sec:ref_eos}
Inspired by the good performance of CNNs in image recognition tasks, we constructed a VGG-like~\cite{simonyan2014very} CNN with 3 convolutional layers followed by fully-connected layers for the EoS binary classification task. The histogram of the pions at min-rapidity is taken as the image-like input for the CNN. Batch normalization, dropout, and PReLU activation are used to avoid overfitting. Details of the network architecture and training are discussed in Ref.\cite{Du:2019civ}. We also refer to Ref.\cite{Pang:2016vdc} for more technical explanations.

\subsection{Cosmic-ray induced Air Showers}\label{sec:ref_airshowers}

To process the time-and-space dependent air-shower footprints, the model uses a two-fold architecture, inspired by Ref.~\cite{ERDMANN201846,pao_dl_2021}.
The first recurrent network part is used to analyze the signal traces recorded by each detector station. Therefore, a two-layer subnetwork  --- consisting of Long short-term memory networks~\cite{hochreiter_long_1997} --- is shared over each of the $9 \times 9$ detector stations and extracts ten features out of each signal trace.

As the resulting output is image-like ($9 \times 9 \times 10$), in the second part, the obtained ten feature maps are concatenated to the map of arrival times ($9 \times 9 \times 1$) and analyzed in the following on spatial correlations using convolutional operations.
This is performed using two blocks, which are separated by a pooling operation. Each of these blocks holds five residual units~\cite{he_deep_2015} with convolutional layers and batch normalization~\cite{ioffe_batch_2015}.
After the residual block, global max pooling and dropout are applied.
To speed up the model's convergence we use a layer that re-normalizes the yielded outputs, which lie at the start of the training mostly between 0 and 1, to the scale of the \xmax distribution.
For the reference architecture, we preprocessed the data by performing a logarithmic re-scaling of the signal traces to compensate for the very different signal sizes. We further normalized the arrival times with respect to the time measured at the central station (station with the largest signal). For more details, refer to Ref.~\cite{ERDMANN201846}.

\end{appendix}

\bibliographystyle{bib_style}
\bibliography{4_BIB.bib}   

%
%

\end{document}